\documentclass[
]{ceurart}

\usepackage{listings}
\usepackage{bm}
\usepackage{amsmath}
\usepackage{amssymb}
\usepackage{threeparttable}
\usepackage{tabularx}
\usepackage{color}
\usepackage{booktabs}
\usepackage{verbatim}
\usepackage{amsopn}
\usepackage{amsmath}
\usepackage{multirow}
\usepackage{colortbl}
\usepackage{makecell} 

\usepackage{eucal}
\usepackage{amsfonts}
\usepackage{bbm}
\usepackage{bm}

\usepackage{xspace}

\makeatletter
\DeclareRobustCommand\onedot{\futurelet\@let@token\@onedot}
\def\@onedot{\ifx\@let@token.\else.\null\fi\xspace}

\def\eg{\emph{e.g}\onedot} 
\def\ie{\emph{i.e}\onedot}

\makeatother

\begin{document}

\copyrightyear{2024}
\copyrightclause{Copyright for this paper by its authors.
  Use permitted under Creative Commons License Attribution 4.0
  International (CC BY 4.0).}

\conference{The 3rd Vision-based Remote Physiological Signal Sensing (RePSS) Challenge \& Workshop, Aug 3--9, 2024, Jeju, South Korea}

\title{Joint Spatial-Temporal Modeling and Contrastive Learning for Self-supervised Heart Rate Measurement}

\author[1]{Wei Qian}[%
orcid=0009-0007-9467-6296,
email=qianwei.hfut@gmail.com,
]
\fnmark[1]

\author[3,4]{Qi Li}[%
orcid=0000-0002-8655-5781,
email=liqi@stu.ahu.edu.cn,
]
\fnmark[1]

\author[5]{Kun Li}[%
orcid=0000-0001-5083-2145,
email=kunli.hfut@gmail.com,
]
\cormark[1]

\author[4,3]{Xinke Wang}[%
orcid=0009-0002-8399-8322,
email=xinkewang689@gmail.com,
]

\author[1,2,3]{Xiao Sun}[%
orcid=0000-0001-9750-7032 ,
email=sunx@hfut.edu.cn,
]

\author[1,2,3]{Meng Wang}[%
orcid=0000-0002-3094-7735,
email=eric.mengwang@gmail.com,
]

\author[1,2,3,6]{Dan Guo}[%
orcid=0000-0003-2594-254X,
email=guodan@hfut.edu.cn,
]
\cormark[1]

\address[1]{School of Computer Science and Information Engineering, School of Artificial Intelligence, Hefei University of Technology (HFUT)}
\address[2]{Key Laboratory of Knowledge Engineering with Big Data (HFUT), Ministry of Education}
\address[3]{Institute of Artificial Intelligence, Hefei Comprehensive National Science Center, China}
\address[4]{Anhui University, China}
\address[5]{Zhejiang University, China}
\address[6]{Anhui Zhonghuitong Technology Co., Ltd.}

\cortext[1]{Corresponding authors.}
\fntext[1]{These authors contributed equally.}

\begin{abstract}
This paper briefly introduces the solutions developed by our team, HFUT-VUT, for Track 1 of self-supervised heart rate measurement in the 3rd Vision-based Remote Physiological Signal Sensing (RePSS) Challenge hosted at IJCAI 2024. 
The goal is to develop a self-supervised learning algorithm for heart rate (HR) estimation using unlabeled facial videos. 
To tackle this task, we present two self-supervised HR estimation solutions that integrate spatial-temporal modeling and contrastive learning, respectively. 
Specifically, we first propose a non-end-to-end self-supervised HR measurement framework based on spatial-temporal modeling, which can effectively capture subtle rPPG clues and leverage the inherent bandwidth and periodicity characteristics of rPPG to constrain the model. 
Meanwhile, we employ an excellent end-to-end solution based on contrastive learning, aiming to generalize across different scenarios from complementary perspectives. 
Finally, we combine the strengths of the above solutions through an ensemble strategy to generate the final predictions, leading to a more accurate HR estimation. 
As a result, our solutions achieved a remarkable RMSE score of 8.85277 on the test dataset, securing \textbf{2nd place} in Track 1 of the challenge.
\end{abstract}


\maketitle

\section{Introduction}
Remote physiological measurement~\cite{li2014remote,li2018obf,qian2024dual,li2023channel,liu2023efficientphys} has emerged as a promising field with significant applications in healthcare, wellness monitoring, and human-computer interaction. Traditional methods for physiological measurement, such as electrocardiograms (ECG) and photoplethysmograms (PPG), require direct contact with the skin, which can be cumbersome and inconvenient for continuous monitoring. 
With the great success of deep learning in computer vision~\cite{tang2022gloss,zhou2022contrastive,li2023vigt,guo2024benchmarking,wei2021deraincyclegan}, recent advancements~\cite{sun2024contrastphys,lu2021dual} have paved the way for non-contact, video-based techniques to estimate physiological signals such as heart rate (HR) and respiratory rate (RR) from facial videos, providing a more comfortable and accessible approach for users.

Despite the promising potential of video-based physiological measurement, most existing methods~\cite{yu2019remote,liu2023efficientphys,qian2024dual} rely heavily on supervised learning, necessitating large amounts of labeled data for training. Acquiring such labeled data is often labor-intensive and time-consuming, posing a significant bottleneck for developing robust and generalizable models. Moreover, supervised methods may not generalize well across different environments and lighting conditions, limiting their practical applicability. Therefore, the development of label-free rPPG estimation methods is becoming increasingly urgent.

To address these challenges, the 3rd Vision-based Remote Physiological Signal Sensing (RePSS) Challenge at IJCAI 2024 was launched. 
This challenge aims to develop self-supervised training methods for HR measurement using unlabeled facial videos, thereby reducing the dependency on extensive labeled datasets. 
For this challenge, we present two self-supervised HR estimation solutions that integrate spatial-temporal modeling and contrastive learning, respectively. Inspired by Dual-TL~\cite{qian2024dual} and SiNC~\cite{speth2023non}, we propose a non-end-to-end self-supervised HR measurement framework based on a spatial-temporal Transformer to capture subtle rPPG clues. 
Meanwhile, we adopt a complementary end-to-end contrastive learning solution based on Contrast-Phys+~\cite{sun2024contrastphys} to enhance the model accuracy. Finally, we combine the strengths of both solutions through an ensemble strategy to generate the final predictions, securing second place with the RMSE score of 8.85277. 

In conclusion, the main contributions can be summarized as follows: 
\begin{itemize}
\item We propose a non-end-to-end self-supervised solution based on spatial-temporal modeling. By considering the priors of periodicity consistency and bandwidth limitation of the rPPG signal, we introduce four loss functions to supervise the model effectively. 
\item We present an end-to-end solution based on contrastive learning, which utilizes 3DCNN to extract features and employs a contrastive loss to learn discriminative representations for periodic rPPG signal modeling. 
\item Our solution achieved second place with the RMSE score of 8.85277 on the test dataset in Track 1 of the 3rd Vision-based Remote Physiological Signal Sensing Challenge. The experimental results demonstrate the effectiveness and robustness of our proposed solutions.
\end{itemize}

\section{Methodology}
\subsection{Solution 1: Self-supervised HR Measurement with Spatial-Temporal Transformer} 
\label{sec:solution1}
\begin{figure*}[t]
\centering
\includegraphics[width=1.0\linewidth]{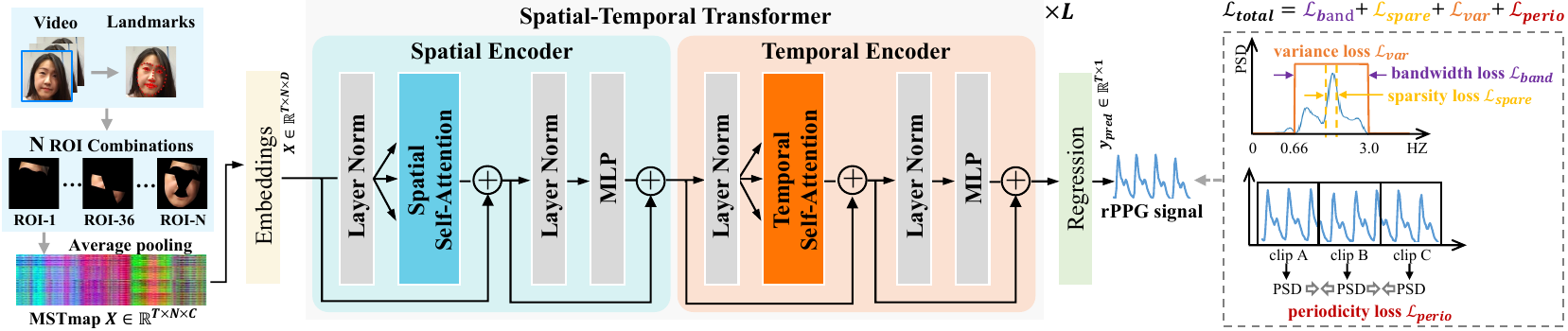}
\caption{Overview of the proposed solution 1. Given an input facial video with $T$ frames, we obtain $N$ facial ROIs for each frame and extract the MSTmap representation $M \in \mathbb{R}^{T\times N\times C}$ for the video, where $N$ is the number of facial ROI. A feature embedding layer is used to project the MSTmap to high-dimensional feature $X \in \mathbb{R}^{T \times N \times D}$. 
Then, we stack spatial-temporal Transformer for $L$ loops to capture subtle rPPG clues. Next, a rPPG regression head is used to output rPPG signal $s_{pre} \in \mathbb{R}^{T \times 1}$. Finally, we apply four self-supervised losses to constrain the model.
}
\label{fig:pipeline}
\end{figure*}

Inspired by the great success of Transformer in computer vision~\cite{li2023transformer}, we present a non-end-to-end self-supervised HR measurement framework to mitigate the need for labeled video data based on a Spatial-Temporal Transformer. 
The overview of this solution is illustrated in Figure~\ref{fig:pipeline}. Specifically, we first transform the input facial video into a multi-scale spatial-temporal map (MSTmap) in Section~\ref{sec:pre-processing}. Then, we introduce our spatial-temporal Transformer module in Section~\ref{sec:STFormer}. Next, in Section~\ref{sec:self-supervisedloss}, with the constraints of periodicity consistency and bandwidth finiteness, our model directly discovers blood volume pulses from unlabeled videos to predict HR.

\subsubsection{Data Pre-processing}~\label{sec:pre-processing}
The quasi-periodic pulse signal originates from subtle light reflections of blood vessels under the skin. Therefore, non-skin pixels and facial geometric features can be considered as rPPG-independent noises. We transform the raw facial video into MSTmap to highlight the spatiotemporal information of the human face, which is a common practice in rPPG measurement~\cite{niu2019rhythmnet,niu2020video}. 
Concretely, the MSTmap divides the facial area into $6$ meta-ROI blocks, which can generate $N$ = ($2^6$-1)=63 ROI combination blocks, and the pixels of each block are averaged separately for $C$ color channels. In the video, all the frames are concatenated along the time dimension to generate a spatial-temporal map of size $\mathbb{R}^{T \times N \times C}$, where $C$ = 6 represents \{R,G,B,Y,U,V\} channels. Next, we embed the MSTmap $M$ to high-dimensional feature $\bm{X} \in \mathbb{R}^{T \times N \times D}$ with feature dimension $D$ by using a full-connected layer.

\subsubsection{Spatial-Temporal Transformer}\label{sec:STFormer}
Our spatial-temporal Transformer tailored for remote physiological measurement is designed carefully for perceiving the temporal and spatial correlations. It includes two encoders (spatial encoder and temporal encoder) to refine the ROI representation containing rPPG clues by capturing long-term spatiotemporal contextual information. We now explain the proposed model in detail. 
Specifically, given the input features $\bm{X} \in \mathbb{R}^{T\times N\times D}$, the process of embedding spatial context for $t$-frame can be formulated as:
\begin{equation}
\begin{aligned}
&{\bm{Q}^{(t)}}=\bm{X}^{(t)} {W_{tq}}, {\bm{K}^{(t)}}=\bm{X}^{(t)}W_{tk}, {\bm{V}^{(t)}}=\bm{X}^{(t)}W_{tv},\\
&{\bm{Z}^{(t)}}=\mathrm{softmax}(\frac{\bm{Q}^{(t)}  {\bm{K}^{(t)}}^T}{\sqrt{D}})\bm{V}^{(t)} + \bm{X}^{(t)},\\
&{\bm{Z}^{'(t)}}=\mathrm{MLP}(\mathrm{LN}(\bm{Z}^{(t)}))+{\bm{Z}^{(t)}},
\end{aligned}
\label{eq:spatia_Transformer}
\end{equation}
where $W_{tq}, W_{tk}, W_{tv}$ are learnable parameters shaped as $D \times D$. $\bm{X}^{(t)}$ denote the feature in $t$-th frame. 
$\mathrm{MLP}$ is the multi-layer perceptron layer and $\mathrm{LN}$ is layer normalization operation. The feature map of all frames $\{\bm{Z}^{'(t)}|t=1,\ldots,T\}$ are concatenated together into $\bm{Z}_{s} \in \mathbb{R}^{T \times N \times D}$.

The other complementary module is applied to enhance the input rPPG features with temporal dynamical transition clues and enrich the temporal context by highlighting the informative features along the time dimension for each facial ROI. 
Our temporal encoder follows the way in Eq.~\ref{eq:spatia_Transformer}. 
The difference is that we calculate the temporal dimension for each spatial unit ($n\in [1, N]$). 
We output the temporally correlated feature for the $n$-th facial ROI feature as $\bm{Z}^{'(n)} \in \mathbb{R}^{T \times D}$ and stack the features $\{ \bm{Z}^{'(n)} | n=1,2,\ldots,N\} $ together, represented by $\bm{Z}_{t} \in \mathbb{R}^{N \times T \times D}$. 

The spatial and temporal encoders are stacked as $L$ loops in an alternating manner, taking into account the spatial and temporal complementary contextual information integrally. Moreover, spatial and temporal position embedding is applied only to the first encoder to retain two kinds of position information. Finally, we use an rPPG regression head to project the feature to a 1D rPPG signal $y_{pred} \in \mathbb{R}^{T \times 1}$.

\subsubsection{Self-supervised Loss}\label{sec:self-supervisedloss}
As highlighted in previous studies~\cite{gideon2021way,speth2023non}, the rPPG signal possesses inherent theoretical priors, including specific bandwidth in the frequency domain. By incorporating this prior knowledge, we employ three self-supervised loss functions from~\cite{speth2023non} in this work. Additionally, to further effectively train the model, we also propose a new periodicity loss based on periodic characteristics of the rPPG signal. Notably, all predicted rPPG signals are transformed into power spectrum density (PSD) with the Fast Fourier Transform (FFT) before computing all losses in our method, denoted as $F = \mathrm{FFT}(y)$. 

\textbf{Bandwidth Loss.}
A healthy HR falls within a specific frequency range. 
Following the~\cite{speth2023non}, we penalize the model for producing signals that exceed the healthy HR bandwidth limits. Consequently, the bandwidth loss can be formalized as follows:
\begin{equation}\label{eq:lb}
\mathcal{L}_{band} = \dfrac{1}{\sum\limits_{i=-\infty}^\infty F_i} \left [ \sum\limits_{i=-\infty}^a F_i + \sum\limits_{i=b}^{\infty} F_i \right ] ,
\end{equation}
where $a$ and $b$ denote lower and upper band limits, respectively. $F_i$ is the power in the $i$th frequency bin of the predicted signal. In our experiments, we specify the limits as $a=0.\overline{66}$ Hz to $b=3$ Hz, which corresponds to a common pulse rate range from 40 bpm to 180 bpm. 
This range effectively captures the typical variations in a healthy HR, ensuring that our model focuses on the relevant frequency components while minimizing the influence of noise. 
By incorporating this bandwidth loss, our model is better equipped to distinguish between meaningful rPPG signals and disturbances, ultimately leading to more accurate HR estimation.

\textbf{Sparsity Loss.}
Since we are primarily interested in heartbeat frequency, we emphasize the periodic heartbeats by suppressing non-heartbeat frequencies. Following~\cite{speth2023non}, we penalize the energy in the bandwidth regions far away from the spectrum peak, which can ensure that the model focuses on the relevant heartbeat frequencies. It can be formulated as:
\begin{equation}\label{eq:ls}
\mathcal{L}_{sparse} = \dfrac{1}{\sum\limits_{i=a}^b F_i} \left [ \sum\limits_{i=a}^{\mathrm{argmax}(F) - \Delta_F} F_i + \sum\limits_{i=\mathrm{argmax}(F) + \Delta_F}^b F_i \right ] ,
\end{equation}
where $\mathrm{argmax}(F)$ is the frequency of the spectral peak, and $\Delta_F=6$ is the frequency padding around the peak. 
This loss enhances the model's ability to accurately estimate HR by ensuring that the spectral energy is concentrated around the true HR frequencies, thus minimizing the influence of noise and other non-relevant frequency components.

\textbf{Variance Loss.}
To avoid the model collapsing to a specific frequency, we also use a variance loss~\cite{speth2023non,bardes2022vicreg} to spread the variance of the power spectral density into a uniform distribution over the desired frequency band. Firstly, we define a uniform prior distribution $P$ over $d$ frequencies. Then, we consider a batch of $n$ spectral densities, represented as $F = [v_1, \ldots, v_n]$, where each $v_i$ is a $d$-dimensional frequency decomposition of a predicted waveform. To aggregate these spectral densities, we compute the normalized sum across the batch, denoted as $Q$. Therefore, the variance loss $\mathcal{L}_{var}$ can be formulated as:
\begin{equation}
\mathcal{L}_{var} = \frac{1}{d} \sum_{i=1}^d \left( \mathrm{CDF}_i(Q) - \mathrm{CDF}_i(P) \right)^2,
\end{equation}
where $\mathrm{CDF}_i$ represents the cumulative distribution function at the $i$-th frequency.

\textbf{Periodicity Loss.}
In addition to the intrinsic properties of the rPPG signal itself, we have observed that adjacent rPPG signals do not change rapidly over short periods. This is typically manifested by similar periodicity in neighboring rPPG signals, meaning they share a dominant peak in the PSD. Specifically, we uniformly sample $S$ non-overlapping temporal segments from a short rPPG signal (\eg, 10s). The PSDs of these segments should be similar. Thus, our proposed periodicity loss can be formulated as:
\begin{equation}
\mathcal{L}_{perio} = \sum_{j=1}^{S-1} \sum_{i=-\infty }^{\infty} \left ( F_i^j -F_i^{j+1} \right ) ^{2},
\end{equation}
where $S=3$ denotes the number of segments. 

In summary, the overall loss function of our self-supervised learning strategy is :
\begin{equation}
\mathcal{L}_{total} = \mathcal{L}_{band} + \mathcal{L}_{sparse} + \mathcal{L}_{var} + \mathcal{L}_{perio}.
\end{equation}

\begin{figure*}[t]
\centering
\includegraphics[width=1.0\linewidth]{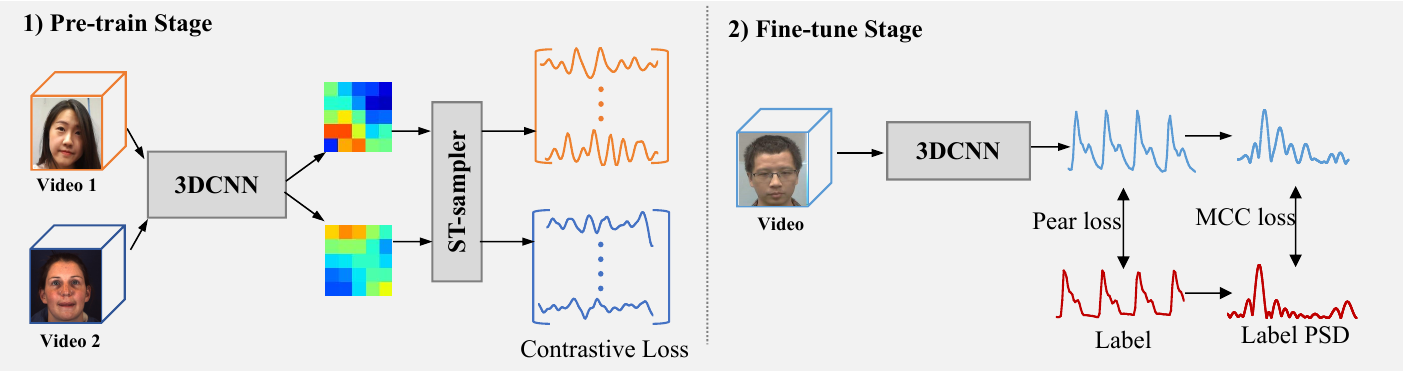}
\caption{Overview of the solution 2. In the pre-train stage, the model is trained in a contrastive learning-based self-supervised manner. After that, the pre-trained model is fine-tuned by supervised loss.}
\label{fig:pipeline2}
\end{figure*}

\subsection{Solution 2: Self-supervised HR Measurement with Contrastive Learning} \label{sec:solution2}
Here we provide the end-to-end self-supervised HR measurement framework based on the contrastive learning strategy. The framework is depicted in Figure~\ref{fig:pipeline2}. Specifically, we first perform data-preprocessing in Section\ref{s2:datapro}. Then we pre-train the proposed model in an unsupervised setting based on the Contrast-Phys+~\cite{sun2024contrastphys} in Section~\ref{s2:pretrain}. Finally, we fine-tune the Contrast-Phys+ model with a supervised setting and obtain the final rPPG predictor in Section~\ref{s2:finetune}.

\subsubsection{Data Pre-processing}\label{s2:datapro}
In this self-supervised manner, we input facial video into our model to estimate the final rPPG signal. For an original video, we first perform face detection by MTCNN~\cite{zhang2016_mtcnn} to get the four coordinates of the face bounding box from the first frame. Then, we enlarge the length and width of the bounding box by 1.5 times and crop the face region for each frame of the video. The cropped faces are resized to $128 \times 128$. Next, we segment each video into clips to feed into the model. Note that we also perform frame difference operations on the clip to generate normalized difference frames as an attempt of model input. 
The difference between two consecutive frames can be formulated as:
\begin{equation}
\Delta V_{t} = V_{t+1} - V_{t},
\end{equation}
where $V_{t}$ denotes the $t$-th frame. To keep the length of the difference video equal to the raw video, we simply repeat the last difference frame. 
Then, the $\Delta V$ is normalized.

\subsubsection{Pre-training}\label{s2:pretrain}
In this stage, following the setting of~\cite{sun2024contrastphys} we modify the 3DCNN-based PhysNet to get spatiotemporal rPPG (ST-rPPG) block representation. The model outputs spatiotemporal rPPG features with shape $T\times S\times S$, where $T$ is the temporal length, and $S$ is the spatial dimension. 
The ST-rPPG block can be regarded as a collection of rPPG signals from different facial regions. 
Therefore, for each input, we can sample $S^2$ rPPG signals with the length of $T$. 

According to the observations that rPPG spatial similarity and temporal similarity in~\cite{sun2024contrastphys}, the ST-rPPG block can sample multiple rPPG signals with short time intervals and different spatial positions. Those signals should be similar. Then contrastive learning can be formulated by pulling together the rPPG signals from the same ST-rPPG block and pushing away the signals from different ST-rPPG blocks extracted in the crossing video. The contrastive loss can be formulated as:
\begin{align}
&\mathcal{L}_{pos} = \sum_{i=1}^{N} \sum_{\substack{j=1 \\ j \neq i}}^{N}\left(\left\|f_{i}-f_{j}\right\|^{2}+\left\|f_{i}^{\prime}-f_{j}^{\prime}\right\|^{2}\right) /(2 N(N-1)), \\
&\mathcal{L}_{neg} = -\sum_{i=1}^{N} \sum_{j=1}^{N}\left\|f_{i}-f_{j}^{\prime}\right\|^{2} / N^{2},\\
    &\mathcal{L}_{ctr} = \mathcal{L}_{pos} + \mathcal{L}_{neg} ,
\end{align}
where $f_i$ denotes the Power Spectrum Densities (PSDs) of the rPPG signal in position $i$ and $f_{i}^{\prime}$ is the other video's PSDs. $N$ is the number of sampled rPPG pairs. 
The contrastive loss function minimizes the MSE distance between positive samples and maximizes the distance between the negative samples to force the model to learn the discriminative representation of the underlying signals from different videos.

\subsubsection{Fine-tuning}\label{s2:finetune}
With the pre-trained 3DCNN-based PhysNet model, we use the officially designated dataset to fine-tune it in a supervised manner. 
Specifically, in this stage, we modified the output of the model by averaging the spatial dimension and then obtained a predicted rPPG signal. Given the predicted rPPG signal $y_{pred}$ and the ground-truth PPG signal $y_{gt}$, a popular Negative Pearson correlation (Pear) loss and Negative max cross-correlation (MCC) loss are selected to perform supervised training. It is worth noting that the Pear is the time domain loss function while the MCC loss is the frequency domain loss function. The MCC is robust to temporal offsets in the ground truth, which can make up for the Pear loss. The MCC loss is formulated as:
\begin{equation}
\mathcal{L}_{mcc}= - \operatorname{Max}\left(\frac{FFT^{-1}\{BPass(FFT\{y_{pred}\} \cdot \overline{FFT\{y_{gt}\}})}{\sigma_{y_{pred}} \times \sigma_{y_{gt}}}\right),
\end{equation}
where $FFT^{-1}$ is the inverse of fast Fourier transform (FFT), $\sigma$ is the standard deviation. 
Besides, as the ground-truth signals are the reference of predicted rPPG signals, the $y_{pred}$ should be similar to $y_{gt}$. Therefore, we also use the contrastive loss by the following:
\begin{align}
&\mathcal{L}^{gt}_{pos} = \sum_{i=1}^{N} \sum_{\substack{j=1 \\ j \neq i}}^{N}\left(\left\|f_{i}-g_{j}\right\|^{2}+\left\|f_{i}^{\prime}-g_{j}^{\prime}\right\|^{2}\right) /(2 N(N-1)), \\
&\mathcal{L}^{gt}_{neg} = -\sum_{i=1}^{N} \sum_{j=1}^{N}\left(\left\|f_{i}-g_{j}^{\prime}\right\|^{2}+\left\|f_{i}^{\prime}-g_{j}\right\|^{2} \right) / N^{2},
\end{align}
where $g$ is the PSDs of the ground-truth signal.

Finally, the overall loss for fine-tuning is the combination of Pear loss, MCC loss, and contrastive loss, which can resist noise interference of ground-truth signal. 
\begin{equation}
\mathcal{L}_{s} = \mathcal{L}^{gt}_{pos} + \mathcal{L}^{gt}_{neg} + \alpha \mathcal{L}_{pear} + \beta \mathcal{L}_{mcc},
\end{equation}
where $\mathcal{L}_{pear}$ is the Negative Pearson correlation loss function. In our experiments, we set $\alpha$ to 0.1 and $\beta$ to 0.2 for the VIPL-V2 dataset.

\section{Experiments}
\subsection{Datasets}
\textbf{UBFC-rPPG}~\cite{bobbia2019unsupervised} is a commonly used pure dataset for physiological estimation. It records 42 facial videos from 42 subjects in a stable lab environment. 
\textbf{PURE}~\cite{stricker2014non} contains 60 facial videos of 10 participants under 6 modes (steady, small rotation, medium rotation, talking, slow translation, and fast translation). 
\textbf{MMSE-HR}~\cite{tulyakov2016self} contains 102 facial videos captured from 40 subjects under six task modes. This dataset contains various facial expression changes. 
\textbf{DISFA}~\cite{DISFA} is a non-posed facial expression dataset. It records 27 facial videos from 27 subjects with different ethnicities\cite{DISFA-2}. 
\textbf{VIPL-V2}~\cite{niu2019vipl} is the second version of the VIPL-HR~\cite{niu2019vipl} dataset for remote HR estimation from face videos under less-constrained situations, which contains 2,000 RGB videos provided in this challenge~\cite{niu2019rhythmnet,niu2020video}. 
Up until the publication of the \textbf{OBF}~\cite{li2018obf} dataset, it contains 100 healthy subjects and 6 patients with atrial fibrillation, totaling 10,600 minutes in length~\cite{yu2019remote}. 
In this challenge, some data of OBF are included in the test set.
Following the rule of this challenge, we use the datasets except VIPL-V2 and OBF without labels to pre-train the model and finetune the model on the VIPL-V2 dataset.

\subsection{Evaluation Metrics and Implementation Details}
In this challenge, the root mean squared error (RMSE) is selected as the evaluation metric between the predicted HR $y_{pred}$ and ground-truth HR $y_{gt}$ as below:
\begin{equation}
RMSE(y_{pred}, y_{gt}) = \sqrt{\frac{1}{N} {\textstyle \sum_{i=1}^{N}} (y_{pred}^{i}-y_{gt}^{i})},
\end{equation}
where $N$ denotes the number of video samples. 

For solution 1 introduced in Section~\ref{sec:solution1}, we begin by extracting the facial ROI regions using the landmark detection tool of OpenFace during the data pre-processing step. We then follow the setting described in~\cite{niu2020video}, applying a sliding window size of 300 frames (10s) and a step size of 15 frames (0.5s) to generate MSTmap from the facial videos. 
For the spatial-temporal Transformer module, we set the dimensionality $D$ to 128 and the number of layers $L$ to 6. 
During pre-training, we use the AdamW optimizer with a learning rate of 1e-4 and a batch size of 4. 
Data augmentation techniques include random horizontal and vertical flipping as well as frequency up/down sampling are used. 
In the fine-tuning step with data labels, in addition to the self-supervised loss, we also add Negative Pearson Loss to further optimize the model. 
Besides, we use a smaller learning rate, \ie, 1e-5, to finetune the model.
For the VIPL-V2 dataset, we split the training and validation subsets in a ratio of 8:2. 
For the HR estimation inference step, following previous work~\cite{qian2024dual,li2023channel}, we apply a 1st-order Butterworth filter to convert the rPPG signal into an HR value with a cutoff frequency range of [0.66Hz, 3.0Hz], corresponding to [40, 180] beats per minute. Subsequently, we perform the PSD~\cite{psd} to estimate HR for each video clip. 
For solution 2 elaborated in Section~\ref{sec:solution2}, we resample the videos to a frame rate of 30 and then perform face detection and cropping. 
We set the length of the video clip to 300 frames without overlapping.  
Following the setting in~\cite{sun2024contrastphys}, the spatial resolution $S$ is set to 2, and the sampled time interval $\Delta t$ of each rPPG signal is set to 150 frames. Other settings are the same as solution 1.

For the ensemble strategy, we take the multiple best prediction results under different settings of both solution 1 and solution 2. Then we average the different predicted heart rates of each sample as the final result.

\begin{table}[t]
\centering
\caption{The ablation study results of our solution 1 on the test dataset.}
\resizebox{0.8\linewidth}{!}{
\begin{tabular}{lll|c}
\hline
Pre-training & Fine-tuning & Loss & RMSE$\downarrow$ (bpm)     \\ \hline
\multirow{2}{*}{UBFC-rPPG}   & \multirow{2}{*}{VIPL-V2} & $\mathcal{L}_{band} + \mathcal{L}_{sparse} + \mathcal{L}_{var}$ & 13.88440 \\
 &  & $\mathcal{L}_{band} + \mathcal{L}_{sparse} + \mathcal{L}_{var} + \mathcal{L}_{perio}$ & 12.30601 \\ \hline
 \multirow{2}{*}{UBFC-rPPG + PURE} & \multirow{2}{*}{VIPL-V2} & $\mathcal{L}_{band} + \mathcal{L}_{sparse} + \mathcal{L}_{var}$ & 11.52003 \\
 &  & $\mathcal{L}_{band} + \mathcal{L}_{sparse} + \mathcal{L}_{var} + \mathcal{L}_{perio}$ & 10.67180 \\ \hline
 \multirow{2}{*}{UBFC-rPPG + PURE + MMSE-HR} & \multirow{2}{*}{VIPL-V2} & $\mathcal{L}_{band} + \mathcal{L}_{sparse} + \mathcal{L}_{var}$ & 10.36720 \\
 &  & $\mathcal{L}_{band} + \mathcal{L}_{sparse} + \mathcal{L}_{var} + \mathcal{L}_{perio}$ & 9.93125 \\
\hline
\end{tabular}}
\label{tab:solution1_result}
\end{table}

\subsection{Experimental Results}
\noindent\textbf{Results for Solution 1.} 
As shown in Table~\ref{tab:solution1_result}, we investigate the impact of different pre-training datasets and loss functions for solution 1. The results indicate that as the amount of pre-training data increases, the performance of the model improves accordingly. 
In our solution, we ultimately select the UBFC-rPPG~\cite{bobbia2019unsupervised}, PURE~\cite{stricker2014non}, and MMSE-HR~\cite{tulyakov2016self} datasets for pre-training. 
Additionally, we also investigate the impact of the proposed periodicity loss $\mathcal{L}_{perio}$. 
We can see that the incorporation of the periodicity loss consistently improves the performance of the model significantly across different settings. For instance, when the model is trained on the UBFC-rPPG, PURE, and MMSE-HR datasets, the introduction of the periodicity loss reduces RMSE from 10.35720 to 9.93125. 
This improvement underscores the effectiveness of the periodicity loss in mitigating abnormal periodic fluctuations in the predicted signal and maintaining temporal periodicity consistency.

\begin{table}[t]
\centering
\caption{The ablation study results of our solution 2 on the test dataset. * denotes the normalized difference on model input.}
\resizebox{\linewidth}{!}{
\begin{tabular}{lll|c}
\hline
Pre-training & Fine-tuning & Loss & RMSE$\downarrow$ (bpm)     \\ \hline
\multirow{3}{*}{DISFA} & \multirow{3}{*}{VIPL-V2} & $\mathcal{L}^{gt}_{pos} + \mathcal{L}^{gt}_{neg}$ & 11.81139 \\
 &  & $\mathcal{L}^{gt}_{pos} + \mathcal{L}^{gt}_{neg} + \alpha\mathcal{L}_{pear}$ & 12.01150 \\
 &  & $\mathcal{L}^{gt}_{pos} + \mathcal{L}^{gt}_{neg} + \beta\mathcal{L}_{mcc}$ & 11.29330 \\
 \hline
\multirow{2}{*}{DISFA + MMSE-HR} & \multirow{2}{*}{VIPL-V2} & $\mathcal{L}^{gt}_{pos} + \mathcal{L}^{gt}_{neg}$ & 11.35523 \\
 &  & $\mathcal{L}^{gt}_{pos} + \mathcal{L}^{gt}_{neg} + \alpha\mathcal{L}_{pear} + \beta\mathcal{L}_{mcc}$ & 10.72491 \\
 \hline
 \multirow{3}{*}{DISFA + UBFC-rPPG + MMSE-HR} & \multirow{3}{*}{VIPL-V2} & $\mathcal{L}^{gt}_{pos} + \mathcal{L}^{gt}_{neg}$ & 10.37686 \\
 &  & $\mathcal{L}^{gt}_{pos} + \mathcal{L}^{gt}_{neg} + \beta\mathcal{L}_{mcc}$ & 11.03058 \\
 &  & $\mathcal{L}^{gt}_{pos} + \mathcal{L}^{gt}_{neg} + \alpha\mathcal{L}_{pear} + \beta\mathcal{L}_{mcc}$ & 10.75880 \\
 \hline
 \multirow{3}{*}{DISFA + UBFC-rPPG + MMSE-HR + PURE} & \multirow{3}{*}{VIPL-V2} & $\mathcal{L}^{gt}_{pos} + \mathcal{L}^{gt}_{neg}$ & 10.62485 \\
 &  & $\mathcal{L}^{gt}_{pos} + \mathcal{L}^{gt}_{neg} + \beta\mathcal{L}_{mcc}$ & 10.19808 \\
 &  & $\mathcal{L}^{gt}_{pos} + \mathcal{L}^{gt}_{neg} + \alpha\mathcal{L}_{pear} + \beta\mathcal{L}_{mcc}$ & 11.01228 \\
 \hline
 * DISFA + UBFC-rPPG + MMSE-HR + PURE & VIPL-V2 & $\mathcal{L}^{gt}_{pos} + \mathcal{L}^{gt}_{neg} + \alpha\mathcal{L}_{pear} + \beta\mathcal{L}_{mcc}$ & 10.36316 \\
\hline
\end{tabular}}
\label{tab:solution2_result}
\end{table}

\begin{table}[t]
\centering
\caption{The results of the top-3 leaderboards on the test dataset in each challenge of RePSS. The best result is highlighted in \textbf{bold}, and the second-best result is \underline{underlined}. The results of 1st and 2nd RePSS are provided by the report~\cite{li20201st,li20212nd}, and the 3rd results are provided by the Kaggle competition page\protect\footnotemark[1].}
\resizebox{0.8\linewidth}{!}{
\begin{tabular}{cccc|c}
\hline
Team Name & Venue & Rank & Method Type & RMSE$\downarrow$ (bpm)     \\ \hline
Mixanik   & 1st RePSS & 1 & Supervised & 10.68021 \\
PoWeiHuang & 1st RePSS & 2 & Supervised & 14.16263 \\
AWoyczyk & 1st RePSS & 3 & Supervised & 14.37509 \\ \hline
Dr.L & 2nd RePSS & 1 & Supervised & 11.05 \\
TIME & 2nd RePSS & 2 & Supervised & 11.44 \\
The Anti-Spoofers & 2nd RePSS & 3 & Supervised & 14.51 \\ \hline
Face AI & 3rd RePSS & 1 & Self-supervised & \textbf{8.50693} \\
HFUT-VUT (\textbf{Ours}) & 3rd RePSS & 2 & Self-supervised & \underline{8.85277} \\
PCA\_Vital & 3rd RePSS & 3 & Self-supervised & 8.96941 \\
\hline
\end{tabular}}
\label{tab:leaderboard}
\end{table}
\footnotetext[1]{\href{https://www.kaggle.com/competitions/the-3rd-repss-t1/leaderboard}{https://www.kaggle.com/competitions/the-3rd-repss-t1/leaderboard}
}

\noindent\textbf{Results for Solution 2.}
As shown in Table~\ref{tab:solution2_result}, we evaluate different pre-training datasets, loss functions, and model inputs to find the best setting for this task. 
Note that the DISFA dataset is a non-posed facial expression database. However, from the results, we can find that using it for pre-training can still achieve comparable performance. Apart from that, we can find the same conclusion as solution 1 that increasing the amount of pre-training datasets is beneficial to performance. In this solution, we choose DISFA, UBFC-rPPG, MMSE-HR, and PURE for pre-training. Additionally, we also evaluate different combinations of supervised loss $\mathcal{L}_{s}$. The results show that both the time domain and frequency domain loss are helpful for model fine-tuning. Moreover, we evaluate the performance of normalized frame difference input, and it shows a comparable result with normal input. 
In the model ensemble phase, we added the frame difference-based manner as different feature forms. 

\noindent\textbf{Model Ensemble.} 
In order to combine the advantages of Solution 1 and Solution 2, we use an ensemble strategy to integrate the best prediction results of these two solutions together. 
Specifically, we ensembled the models by taking the average value of the prediction results for Solution 1 and Solution 2, and then obtained the final prediction results.
As shown in Table~\ref{tab:leaderboard}, we report the top-3 results on the test dataset for each RePSS challenge. 
Compared to other teams, we can see that our team achieves 2nd place, which is higher than the 3rd by 1.2\%. This demonstrates that our proposed two self-supervision solutions can complementaryly achieve more accurate and robust heart rate estimation. 
Compared to the results of the supervised methods in previous challenges, we can find that self-supervised methods improve performance by a large margin. 
This indicates that self-supervised methods can capture rPPG-related signals from facial videos during the pre-train phase without requiring any real physiological signals. 

\section{Conclusion}
In this paper, we present our solutions developed for self-supervised remote heart rate measurement of the 3rd RePSS challenge hosted at IJCAI 2024. 
Specifically, we propose two self-supervised HR estimation solutions that integrate spatial-temporal modeling and contrastive learning, respectively. By leveraging the ensemble strategy, our final submission takes second place with the RMSE score of 8.85277 bpm. In the future, we plan to address the issues in this challenge from other perspectives, \eg, using video motion magnification algorithms~\cite{wang2024eulermormer} to capture the subtle change reflected in faces by heartbeats. 

\begin{acknowledgments}
This work was supported by the National Key R\&D Program of China (NO.2022YFB4500601), the National Natural Science Foundation of China (72188101,62272144,62020106007and U20A20183), the Major Project of Anhui Province(202203a05020011), and the Fundamental Research Funds for the Central Universities. 
\end{acknowledgments}

\bibliography{sample-ceur}

\begin{thebibliography}{30}
\expandafter\ifx\csname natexlab\endcsname\relax\def\natexlab#1{#1}\fi
\providecommand{\url}[1]{\texttt{#1}}
\providecommand{\href}[2]{#2}
\providecommand{\path}[1]{#1}
\providecommand{\DOIprefix}{doi:}
\providecommand{\ArXivprefix}{arXiv:}
\providecommand{\URLprefix}{URL: }
\providecommand{\Pubmedprefix}{pmid:}
\providecommand{\doi}[1]{\href{http://dx.doi.org/#1}{\path{#1}}}
\providecommand{\Pubmed}[1]{\href{pmid:#1}{\path{#1}}}
\providecommand{\bibinfo}[2]{#2}
\ifx\xfnm\relax \def\xfnm[#1]{\unskip,\space#1}\fi
\bibitem[{Li et~al.(2014)Li, Chen, Zhao, and Pietikainen}]{li2014remote}
\bibinfo{author}{X.~Li}, \bibinfo{author}{J.~Chen}, \bibinfo{author}{G.~Zhao},
  \bibinfo{author}{M.~Pietikainen},
\newblock \bibinfo{title}{Remote heart rate measurement from face videos under
  realistic situations},
\newblock in: \bibinfo{booktitle}{Proceedings of the IEEE Conference on
  Computer Vision and Pattern Recognition}, \bibinfo{year}{2014}, pp.
  \bibinfo{pages}{4264--4271}.
\bibitem[{Li et~al.(2018)Li, Alikhani, Shi, Seppanen, Junttila, Majamaa-Voltti,
  Tulppo, and Zhao}]{li2018obf}
\bibinfo{author}{X.~Li}, \bibinfo{author}{I.~Alikhani},
  \bibinfo{author}{J.~Shi}, \bibinfo{author}{T.~Seppanen},
  \bibinfo{author}{J.~Junttila}, \bibinfo{author}{K.~Majamaa-Voltti},
  \bibinfo{author}{M.~Tulppo}, \bibinfo{author}{G.~Zhao},
\newblock \bibinfo{title}{The obf database: A large face video database for
  remote physiological signal measurement and atrial fibrillation detection},
\newblock in: \bibinfo{booktitle}{2018 13th IEEE International Conference on
  Automatic Face \& Gesture Recognition (FG 2018)}, \bibinfo{year}{2018}, pp.
  \bibinfo{pages}{242--249}.
\bibitem[{Qian et~al.(2024)Qian, Guo, Li, Zhang, Tian, Yang, and
  Wang}]{qian2024dual}
\bibinfo{author}{W.~Qian}, \bibinfo{author}{D.~Guo}, \bibinfo{author}{K.~Li},
  \bibinfo{author}{X.~Zhang}, \bibinfo{author}{X.~Tian},
  \bibinfo{author}{X.~Yang}, \bibinfo{author}{M.~Wang},
\newblock \bibinfo{title}{Dual-path tokenlearner for remote
  photoplethysmography-based physiological measurement with facial videos},
\newblock \bibinfo{journal}{IEEE Transactions on Computational Social Systems}
  (\bibinfo{year}{2024}).
\bibitem[{Li et~al.(2023)Li, Guo, Qian, Tian, Sun, Zhao, and
  Wang}]{li2023channel}
\bibinfo{author}{Q.~Li}, \bibinfo{author}{D.~Guo}, \bibinfo{author}{W.~Qian},
  \bibinfo{author}{X.~Tian}, \bibinfo{author}{X.~Sun},
  \bibinfo{author}{H.~Zhao}, \bibinfo{author}{M.~Wang},
\newblock \bibinfo{title}{Channel-wise interactive learning for remote heart
  rate estimation from facial video},
\newblock \bibinfo{journal}{IEEE Transactions on Circuits and Systems for Video
  Technology}  (\bibinfo{year}{2023}).
\bibitem[{Liu et~al.(2023)Liu, Hill, Jiang, Patel, and
  McDuff}]{liu2023efficientphys}
\bibinfo{author}{X.~Liu}, \bibinfo{author}{B.~Hill},
  \bibinfo{author}{Z.~Jiang}, \bibinfo{author}{S.~Patel},
  \bibinfo{author}{D.~McDuff},
\newblock \bibinfo{title}{Efficientphys: Enabling simple, fast and accurate
  camera-based cardiac measurement},
\newblock in: \bibinfo{booktitle}{Proceedings of the IEEE/CVF Winter Conference
  on Applications of Computer Vision}, \bibinfo{year}{2023}, pp.
  \bibinfo{pages}{5008--5017}.
\bibitem[{Tang et~al.(2022)Tang, Hong, Guo, and Wang}]{tang2022gloss}
\bibinfo{author}{S.~Tang}, \bibinfo{author}{R.~Hong}, \bibinfo{author}{D.~Guo},
  \bibinfo{author}{M.~Wang},
\newblock \bibinfo{title}{Gloss semantic-enhanced network with online
  back-translation for sign language production},
\newblock in: \bibinfo{booktitle}{Proceedings of the 30th ACM International
  Conference on Multimedia}, \bibinfo{year}{2022}, pp.
  \bibinfo{pages}{5630--5638}.
\bibitem[{Zhou et~al.(2022)Zhou, Guo, and Wang}]{zhou2022contrastive}
\bibinfo{author}{J.~Zhou}, \bibinfo{author}{D.~Guo}, \bibinfo{author}{M.~Wang},
\newblock \bibinfo{title}{Contrastive positive sample propagation along the
  audio-visual event line},
\newblock \bibinfo{journal}{IEEE Transactions on Pattern Analysis and Machine
  Intelligence}  (\bibinfo{year}{2022}).
\bibitem[{Li et~al.(2023)Li, Guo, and Wang}]{li2023vigt}
\bibinfo{author}{K.~Li}, \bibinfo{author}{D.~Guo}, \bibinfo{author}{M.~Wang},
\newblock \bibinfo{title}{Vigt: proposal-free video grounding with a learnable
  token in the transformer},
\newblock \bibinfo{journal}{Science China Information Sciences}
  \bibinfo{volume}{66} (\bibinfo{year}{2023}) \bibinfo{pages}{202102}.
\bibitem[{Guo et~al.(2024)Guo, Li, Hu, Zhang, and Wang}]{guo2024benchmarking}
\bibinfo{author}{D.~Guo}, \bibinfo{author}{K.~Li}, \bibinfo{author}{B.~Hu},
  \bibinfo{author}{Y.~Zhang}, \bibinfo{author}{M.~Wang},
\newblock \bibinfo{title}{Benchmarking micro-action recognition: Dataset,
  methods, and applications},
\newblock \bibinfo{journal}{IEEE Transactions on Circuits and Systems for Video
  Technology}  (\bibinfo{year}{2024}).
\bibitem[{Wei et~al.(2021)Wei, Zhang, Wang, Xu, Yang, Yan, and
  Wang}]{wei2021deraincyclegan}
\bibinfo{author}{Y.~Wei}, \bibinfo{author}{Z.~Zhang},
  \bibinfo{author}{Y.~Wang}, \bibinfo{author}{M.~Xu},
  \bibinfo{author}{Y.~Yang}, \bibinfo{author}{S.~Yan},
  \bibinfo{author}{M.~Wang},
\newblock \bibinfo{title}{Deraincyclegan: Rain attentive cyclegan for single
  image deraining and rainmaking},
\newblock \bibinfo{journal}{IEEE Transactions on Image Processing}
  \bibinfo{volume}{30} (\bibinfo{year}{2021}) \bibinfo{pages}{4788--4801}.
\bibitem[{Sun and Li(2024)}]{sun2024contrastphys}
\bibinfo{author}{Z.~Sun}, \bibinfo{author}{X.~Li},
\newblock \bibinfo{title}{Contrast-phys+: Unsupervised and weakly-supervised
  video-based remote physiological measurement via spatiotemporal contrast},
\newblock \bibinfo{journal}{IEEE Transactions on Pattern Analysis and Machine
  Intelligence}  (\bibinfo{year}{2024}) \bibinfo{pages}{1--18}.
\bibitem[{Lu et~al.(2021)Lu, Han, and Zhou}]{lu2021dual}
\bibinfo{author}{H.~Lu}, \bibinfo{author}{H.~Han}, \bibinfo{author}{S.~K.
  Zhou},
\newblock \bibinfo{title}{Dual-gan: Joint bvp and noise modeling for remote
  physiological measurement},
\newblock in: \bibinfo{booktitle}{Proceedings of the IEEE/CVF Conference on
  Computer Vision and Pattern Recognition}, \bibinfo{year}{2021}, pp.
  \bibinfo{pages}{12404--12413}.
\bibitem[{Yu et~al.(2019)Yu, Peng, Li, Hong, and Zhao}]{yu2019remote}
\bibinfo{author}{Z.~Yu}, \bibinfo{author}{W.~Peng}, \bibinfo{author}{X.~Li},
  \bibinfo{author}{X.~Hong}, \bibinfo{author}{G.~Zhao},
\newblock \bibinfo{title}{Remote heart rate measurement from highly compressed
  facial videos: an end-to-end deep learning solution with video enhancement},
\newblock in: \bibinfo{booktitle}{Proceedings of the IEEE/CVF International
  Conference on Computer Vision}, \bibinfo{year}{2019}, pp.
  \bibinfo{pages}{151--160}.
\bibitem[{Speth et~al.(2023)Speth, Vance, Flynn, and Czajka}]{speth2023non}
\bibinfo{author}{J.~Speth}, \bibinfo{author}{N.~Vance},
  \bibinfo{author}{P.~Flynn}, \bibinfo{author}{A.~Czajka},
\newblock \bibinfo{title}{Non-contrastive unsupervised learning of
  physiological signals from video},
\newblock in: \bibinfo{booktitle}{Proceedings of the IEEE/CVF Conference on
  Computer Vision and Pattern Recognition}, \bibinfo{year}{2023}, pp.
  \bibinfo{pages}{14464--14474}.
\bibitem[{Li et~al.(2023)Li, Li, Guo, Yang, and Wang}]{li2023transformer}
\bibinfo{author}{K.~Li}, \bibinfo{author}{J.~Li}, \bibinfo{author}{D.~Guo},
  \bibinfo{author}{X.~Yang}, \bibinfo{author}{M.~Wang},
\newblock \bibinfo{title}{Transformer-based visual grounding with
  cross-modality interaction},
\newblock \bibinfo{journal}{ACM Transactions on Multimedia Computing,
  Communications and Applications} \bibinfo{volume}{19} (\bibinfo{year}{2023})
  \bibinfo{pages}{1--19}.
\bibitem[{Niu et~al.(2019)Niu, Shan, Han, and Chen}]{niu2019rhythmnet}
\bibinfo{author}{X.~Niu}, \bibinfo{author}{S.~Shan}, \bibinfo{author}{H.~Han},
  \bibinfo{author}{X.~Chen},
\newblock \bibinfo{title}{Rhythmnet: End-to-end heart rate estimation from face
  via spatial-temporal representation},
\newblock \bibinfo{journal}{IEEE Transactions on Image Processing}
  \bibinfo{volume}{29} (\bibinfo{year}{2019}) \bibinfo{pages}{2409--2423}.
\bibitem[{Niu et~al.(2020)Niu, Yu, Han, Li, Shan, and Zhao}]{niu2020video}
\bibinfo{author}{X.~Niu}, \bibinfo{author}{Z.~Yu}, \bibinfo{author}{H.~Han},
  \bibinfo{author}{X.~Li}, \bibinfo{author}{S.~Shan},
  \bibinfo{author}{G.~Zhao},
\newblock \bibinfo{title}{Video-based remote physiological measurement via
  cross-verified feature disentangling},
\newblock in: \bibinfo{booktitle}{Computer Vision--ECCV 2020: 16th European
  Conference, Glasgow, UK, August 23--28, 2020, Proceedings, Part II 16},
  \bibinfo{year}{2020}, pp. \bibinfo{pages}{295--310}.
\bibitem[{Gideon and Stent(2021)}]{gideon2021way}
\bibinfo{author}{J.~Gideon}, \bibinfo{author}{S.~Stent},
\newblock \bibinfo{title}{The way to my heart is through contrastive learning:
  Remote photoplethysmography from unlabelled video},
\newblock in: \bibinfo{booktitle}{Proceedings of the IEEE/CVF International
  Conference on Computer Vision}, \bibinfo{year}{2021}, pp.
  \bibinfo{pages}{3995--4004}.
\bibitem[{Bardes et~al.(2022)Bardes, Ponce, and Lecun}]{bardes2022vicreg}
\bibinfo{author}{A.~Bardes}, \bibinfo{author}{J.~Ponce},
  \bibinfo{author}{Y.~Lecun},
\newblock \bibinfo{title}{Vicreg: Variance-invariance-covariance regularization
  for self-supervised learning},
\newblock in: \bibinfo{booktitle}{International Conference on Learning
  Representations}, \bibinfo{year}{2022}.
\bibitem[{Zhang et~al.(2016)Zhang, Zhang, Li, and Qiao}]{zhang2016_mtcnn}
\bibinfo{author}{K.~Zhang}, \bibinfo{author}{Z.~Zhang},
  \bibinfo{author}{Z.~Li}, \bibinfo{author}{Y.~Qiao},
\newblock \bibinfo{title}{Joint face detection and alignment using multitask
  cascaded convolutional networks},
\newblock \bibinfo{journal}{IEEE Signal Processing Letters}
  \bibinfo{volume}{23} (\bibinfo{year}{2016}) \bibinfo{pages}{1499--1503}.
\bibitem[{Bobbia et~al.(2019)Bobbia, Macwan, Benezeth, Mansouri, and
  Dubois}]{bobbia2019unsupervised}
\bibinfo{author}{S.~Bobbia}, \bibinfo{author}{R.~Macwan},
  \bibinfo{author}{Y.~Benezeth}, \bibinfo{author}{A.~Mansouri},
  \bibinfo{author}{J.~Dubois},
\newblock \bibinfo{title}{Unsupervised skin tissue segmentation for remote
  photoplethysmography},
\newblock \bibinfo{journal}{Pattern Recognition Letters} \bibinfo{volume}{124}
  (\bibinfo{year}{2019}) \bibinfo{pages}{82--90}.
\bibitem[{Stricker et~al.(2014)Stricker, M{\"u}ller, and
  Gross}]{stricker2014non}
\bibinfo{author}{R.~Stricker}, \bibinfo{author}{S.~M{\"u}ller},
  \bibinfo{author}{H.-M. Gross},
\newblock \bibinfo{title}{Non-contact video-based pulse rate measurement on a
  mobile service robot},
\newblock in: \bibinfo{booktitle}{The 23rd IEEE International Symposium on
  Robot and Human Interactive Communication}, \bibinfo{year}{2014}, pp.
  \bibinfo{pages}{1056--1062}.
\bibitem[{Tulyakov et~al.(2016)Tulyakov, Alameda-Pineda, Ricci, Yin, Cohn, and
  Sebe}]{tulyakov2016self}
\bibinfo{author}{S.~Tulyakov}, \bibinfo{author}{X.~Alameda-Pineda},
  \bibinfo{author}{E.~Ricci}, \bibinfo{author}{L.~Yin}, \bibinfo{author}{J.~F.
  Cohn}, \bibinfo{author}{N.~Sebe},
\newblock \bibinfo{title}{Self-adaptive matrix completion for heart rate
  estimation from face videos under realistic conditions},
\newblock in: \bibinfo{booktitle}{Proceedings of the IEEE/CVF Conference on
  Computer Vision and Pattern Recognition}, \bibinfo{year}{2016}, pp.
  \bibinfo{pages}{2396--2404}.
\bibitem[{Mavadati et~al.(2013)Mavadati, Mahoor, Bartlett, Trinh, and
  Cohn}]{DISFA}
\bibinfo{author}{S.~M. Mavadati}, \bibinfo{author}{M.~H. Mahoor},
  \bibinfo{author}{K.~Bartlett}, \bibinfo{author}{P.~Trinh},
  \bibinfo{author}{J.~F. Cohn},
\newblock \bibinfo{title}{Disfa: A spontaneous facial action intensity
  database},
\newblock \bibinfo{journal}{IEEE Transactions on Affective Computing}
  \bibinfo{volume}{4} (\bibinfo{year}{2013}) \bibinfo{pages}{151--160}.
\bibitem[{Mavadati et~al.(2012)Mavadati, Mahoor, Bartlett, and Trinh}]{DISFA-2}
\bibinfo{author}{S.~M. Mavadati}, \bibinfo{author}{M.~H. Mahoor},
  \bibinfo{author}{K.~Bartlett}, \bibinfo{author}{P.~Trinh},
\newblock \bibinfo{title}{Automatic detection of non-posed facial action
  units},
\newblock in: \bibinfo{booktitle}{2012 19th IEEE International Conference on
  Image Processing}, \bibinfo{year}{2012}, pp. \bibinfo{pages}{1817--1820}.
\bibitem[{Niu et~al.(2019)Niu, Han, Shan, and Chen}]{niu2019vipl}
\bibinfo{author}{X.~Niu}, \bibinfo{author}{H.~Han}, \bibinfo{author}{S.~Shan},
  \bibinfo{author}{X.~Chen},
\newblock \bibinfo{title}{Vipl-hr: A multi-modal database for pulse estimation
  from less-constrained face video},
\newblock in: \bibinfo{booktitle}{Computer Vision--ACCV 2018: 14th Asian
  Conference on Computer Vision, Perth, Australia, December 2--6, 2018, Revised
  Selected Papers, Part V 14}, \bibinfo{year}{2019}, pp.
  \bibinfo{pages}{562--576}.
\bibitem[{Welch(1967)}]{psd}
\bibinfo{author}{P.~Welch},
\newblock \bibinfo{title}{The use of fast fourier transform for the estimation
  of power spectra: A method based on time averaging over short, modified
  periodograms},
\newblock \bibinfo{journal}{IEEE Transactions on Audio and Electroacoustics}
  \bibinfo{volume}{15} (\bibinfo{year}{1967}) \bibinfo{pages}{70--73}.
\bibitem[{Li et~al.(2020)Li, Han, Lu, Niu, Yu, Dantcheva, Zhao, and
  Shan}]{li20201st}
\bibinfo{author}{X.~Li}, \bibinfo{author}{H.~Han}, \bibinfo{author}{H.~Lu},
  \bibinfo{author}{X.~Niu}, \bibinfo{author}{Z.~Yu},
  \bibinfo{author}{A.~Dantcheva}, \bibinfo{author}{G.~Zhao},
  \bibinfo{author}{S.~Shan},
\newblock \bibinfo{title}{The 1st challenge on remote physiological signal
  sensing (repss)},
\newblock in: \bibinfo{booktitle}{Proceedings of the IEEE/CVF Conference on
  Computer Vision and Pattern Recognition Workshops}, \bibinfo{year}{2020}, pp.
  \bibinfo{pages}{314--315}.
\bibitem[{Li et~al.(2021)Li, Sun, Sun, Han, Dantcheva, Shan, and
  Zhao}]{li20212nd}
\bibinfo{author}{X.~Li}, \bibinfo{author}{H.~Sun}, \bibinfo{author}{Z.~Sun},
  \bibinfo{author}{H.~Han}, \bibinfo{author}{A.~Dantcheva},
  \bibinfo{author}{S.~Shan}, \bibinfo{author}{G.~Zhao},
\newblock \bibinfo{title}{The 2nd challenge on remote physiological signal
  sensing (repss)},
\newblock in: \bibinfo{booktitle}{Proceedings of the IEEE/CVF International
  Conference on Computer Vision}, \bibinfo{year}{2021}, pp.
  \bibinfo{pages}{2404--2413}.
\bibitem[{Wang et~al.(2024)Wang, Guo, Li, and Wang}]{wang2024eulermormer}
\bibinfo{author}{F.~Wang}, \bibinfo{author}{D.~Guo}, \bibinfo{author}{K.~Li},
  \bibinfo{author}{M.~Wang},
\newblock \bibinfo{title}{Eulermormer: Robust eulerian motion magnification via
  dynamic filtering within transformer},
\newblock in: \bibinfo{booktitle}{Proceedings of the AAAI Conference on
  Artificial Intelligence}, volume~\bibinfo{volume}{38}, \bibinfo{year}{2024},
  pp. \bibinfo{pages}{5345--5353}.

\end{thebibliography}

\end{document}